\documentclass{comsoc2010}
\usepackage{amsmath}
\usepackage{amsthm}

\title{Online Cake Cutting}
\author{Toby Walsh}

\begin{document}
\newtheorem{theorem}{Theorem}
\newtheorem*{example}{Running Example:}
\newtheorem{definition}{Definition}
\newtheorem{myexample}{Example}
\newtheorem{mytheorem}{Theorem}
\newcommand{\myproof}{\noindent {\bf Proof:\ \ }}
\newcommand{\myqed}{\mbox{$\heartsuit$}}


\begin{abstract}
We propose an online form of the 
cake cutting problem. 
This models situations where
players arrive and depart during the
process of dividing a resource. 
We show that well known fair division
procedures like cut-and-choose and
the Dubins-Spanier moving knife procedure can be 
adapted to apply to such online
problems. We propose some
desirable properties that online
cake cutting procedures might possess
like online forms of proportionality
and envy-freeness, and identify
which properties are in fact
possessed by the different online
cake procedures. 
\end{abstract}


\section{Introduction}

\begin{quote}
{\em Congratulations. Today is your birthday so you
take a cake into the office to share with your
colleagues. At tea time, people slowly start to arrive.
However, as some people have to leave early, 
you cannot wait for everyone to arrive
before you start sharing the cake.  How 
do you proceed fairly? }
\end{quote}

This is an example of what we call an {\em online}
cake cutting problem. Most previous studies of
cake cutting procedures have assumed that all the players
are available at the time of the division. Here, 
players arrive and depart (either with their
cake or perhaps after they have eaten their cake) as
the cake is being divided. Such online problems
occur in the real world as in our birthday example, but also 
on the internet where agents are often connecting asynchronously. 

Online cake cutting poses some new challenges. 
On the one hand, the online aspect of such problems makes
fair division more difficult than in
the offline case. How can we 
ensure that a player does not envy
another player when we may have
to distribute cake to the second 
player before the first player
is present (and we can hope to 
determine information about
their valuation function)? On the
other hand, the online aspect of such 
problems may make fair division easier
than in the offline case. If players
don't see cake that has already been
distributed before they arrive, perhaps
they do not envy it?

\section{Online cake cutting}

As is common in the literature \cite{cakecut}, we will
often assume that each player is risk
averse so they maximize
the minimum value of the 
cake that they will receive, regardless of
what the other players do. 
A risk averse player will not choose a strategy that could yield more value
if it also entails the possibility of getting
less value. 
We will also usually assume that each player is
ignorant of the value functions of
the other players. We disuss
relaxing these assumptions in the
conclusions.

We formulate cake cutting 
as dividing the unit interval between
the different players, where each
player has a (typically additive and continuous) valuation 
function on the intervals that they
are allocated.  
We do not suppose that players
assigns the same value to
the whole cake. Although we
can normalize the valuation functions,
we shall see that is not necessary
as all the cake cutting procedures
are {\em scale invariant}. 
Depending on the application,
we may demand that players
receive a continuous slice of cake
or some union of slices.

In an online cake cutting 
problem, the players arrive in some given
order. Players are allocated their cake and then
depart.  The order in which players
depart can be fixed or can be 
change according to how they value
the cake. For example, the player present
who most values a cut slice of cake
might be the next to depart. 
Alternatively, the player to depart might always
be the player who arrived the longest time ago. 
We will assume that at least one
player departs before the last player
arrives (otherwise we can formulate
this as an offline cake cutting procedure). 
To prevent trivial allocations,
we also assume that at least one player
receives some cake. However,
we do not assume that all players
receive cake or that all the cake is 
allocated. 
Formally an online cake cutting problem
is defined by a procedure which given
the valuation functions of the players who
are present in the room and the 
number of players who will take part in total 
either allocates some cake to one of the present players 
(who then departs) or indicates that we 
wait until the next player arrives. 
This can model both a fixed arrival
and departure order, as well as one
in which the order depends on the 
valuation functions. 

An important dimension of online cake
cutting is what is known and by whom
about the total number of players. 
For example, the total number of players
might be known by all players. On the other hand, 
the players might only know a bound on the
total number of players (e.g. you've invited
20 work colleagues to share your birthday cake
but not all of them might turn up).
However, there are several other possibilities
(e.g. certain players might
have complete certainty
about $n$ whilst others have 
complete uncertainty). In addition,
an interesting generalization is when
cake is being allocated before the total
number of players is fixed. 

\section{Desirable properties}

What properties do we want from an
online cake cutting procedure? 
The literature on cake cutting studies
various notions of fainerness like
proportionality and envy freeness, as well
as various forms of
strategy proofness. The generalization
of cake cutting to an online setting gives
rise to some natural extensions of these notions. 

\subsection{Proportionality}

A cake cutting procedure is
\emph{proportional} iff each of the $n$ players assigns
at least $\frac{1}{n}$ of the total value to their piece(s).
Unfortunately, as we shall show, online cake cutting procedures
cannot always be proportional. 
Suppose you only like icing. The problem is
that you may not be able to prevent all the cake
that is iced being distributed before
you enter the room. We therefore consider weaker forms
of proporitionality that are achievable. 
One more limited form of proportionality 
is that any player receives a fair proportion
of the cake that remains when they arrive.
A cake cutting procedure is
\emph{forward proportional} iff 
each player assigns
at least $\frac{r}{n-k}$ of the total value 
of the cake to their pieces
where $r$ is the fraction of the total
value assigned by the player to the
(remaining) cake when they arrive and $k$ is the number of
players who have already left at this point. 

\subsection{Envy freeness}

A stronger notion of fairness is envy freeness. 
A cake cutting procedure is 
\emph{envy free} iff no player values another
player's pieces more than their own.
Note that envy freeness implies proportionality
but not vice versa. 
With online cake cutting, envy freeness
is also impossible to achieve in general. 
We therefore consider weaker forms
of envy freeness that are achievable. 
A cake cutting procedure is
\emph{forward envy free} iff no player values 
the pieces of cake allocated to other players after
their arrival more than their own. 
Players can, however, value the cake allocated to players who
have already departed more than our
own cake.  This models 
situations where, for instance, 
we do not envy cake we don't 
see being allocated, or players eat their cake before departing
and we do not envy cake that has already been eaten.
Note that forward envy freeness implies forward proportionality
but not vice versa. Similarly,
envy freeness implies forward envy freeness but
not vice versa.  
An even weaker form of envy freeness
is when a player does not envy cake that is 
allocated to other players whilst they are in the room. 
A cake cutting procedure is
\emph{immediately envy free} iff no player values 
the pieces of cake allocated to another player after
their arrival and before their departure more than their own. 
Note that forward envy freeness implies immediate envy freeness
but not vice versa. 

\subsection{Equitability}

Another fairness property is equitability. 
A cake cutting procedure is \emph{equitable} iff
all players assign the same value to the pieces of cake
to which they are allocated 
(and so no player envies another’s valuation). 
For 3 or more player, equitability and
envy freeness can be incompatible
\cite{cakecut}. 
Equitability is a difficult property to achieve, even
more so in our online setting. 
Unlike proportionality or envy freeness,
there seems little merit in considering
weaker forms of equitability. Either
all players assign the same value to their
allocated cake or they do not. There is no
advantage to ignoring the value of the
cake allocated to players who have already departed.

\subsection{Efficiency}

Another important notion is efficiency. 
Efficiency is also called Pareto optimality.
A cake cutting procedure is \emph{Pareto optimal} iff 
there is no other allocation to the one returned
that is more valuable for one player
and at least as valuable for the others. 
Note that Pareto optimality does not in itself
ensure fairness since allocating
all the cake to one player is Pareto optimal. 
A cake cutting procedure is \emph{weakly Pareto optimal}
iff there is no
other allocation to the one returned
that is more valuable for all players.
A cake cutting procedure that is Pareto optimal is weakly
Pareto optimal but not vice versa.

\subsection{Strategy proofness}

Another consideration is whether players
have an incentive to act truthfully.
A cake cutting procedure is \emph{weakly truthful} iff 
a player will do at least as well by telling the truth
whatever valuations are held by the other players
\cite{bjkams2006}. 
A stronger notion (often called strategy proofness
in social choice) is that players must not be able to profit 
even when they know how others value the cake. 
As in \cite{clppaaai2010}, we say that
a cake cutting procedure is \emph{truthful} iff 
there are no valuations where a player will do
better by lying. 

\section{Other properties}

We consider some other
properties of (online) cake cutting procedures.

\subsection{Surjectivity}

This property has been
studied in social choice 
but appears to have received
less attention in fair division. 
It indicates whether the cake
can be divided in every possible way. 
A cake cutting procedure is 
\emph{surjective} iff there are 
valuation functions for the players
such that every possible partition of the
cake into $n$ pieces is possible. 
Note that this definition only considers
allocations where each player receives
one continuous slice of cake. 
However, the definition of
surjectivity could be easily extended 
to allocations where players
can receive multiple slices. Our
definition of surjectivity also ignores
which player receives a particular slice. 
If an online cake
cutting procedure is surjective,
then there is an arrival ordering
of the players and valuation functions
such that any given player can
receive a particular slice. 

\subsection{Scale invariance}

Players may have different scales for their valuation
functions. Scale invariance indicates that
this is unimportant. A cake cutting procedure
is \emph{scale invariant} iff 
the allocation of cake is unchanged
when a player's valuation is uniformly
multiplied by a constant factor. 
It turns out that scale invariance is
not difficult to achieve. Indeed, all
the online cake cutting procedures
we shall consider here are scale invariant. 

\subsection{Sequentiality}

In some situations we may want cake to
be cut from one end. This may be the
case, for instance, when the cake represents
time on a shared
device. An onlike cake cutting procedure is \emph{sequential} iff 
the slice given to any player
is to the left of any slice given to a player
who is later to depart. 

\subsection{Order monotonicity}

A player's allocation of cake
typically depends on their
arrival order. We say that 
a cake cutting procedure is
\emph{order monotonic} iff
a player's valuation of their
cake does not decrease when
they are moved earlier in the
arrival ordering (and all other players
have the same arrival ordering). 
Note that as the moved player
can receive cake of greater
value, players who depart
after them may now receive
cake with less value. 
A positive interpretation of order monotonicity
is that players are encouraged
to participate as early as possible.
On the other hand, players who
have to arrive late may
receive less value. 

\section{General results}

The fact that some players may depart
before others arrive place some fundamental
limitations on the fairness of online cake
cutting procedures.

\begin{mytheorem}
No online cake cutting procedure
is proportional, envy free or equitable.
\end{mytheorem}
\myproof
Suppose the procedure is proportional.
Then every player is allocated some
cake. As the cake cutting procedure is online,
at least one player departs before 
the final player arrives.
Since the valuation function
of the final player to arrive is not
known when the first player
departs, the cake allocated to the
first player to depart cannot depend
on the valuation function of
the final player to arrive. Similarly,
the valuation function of the final
player to arrive cannot
change who is the first player to depart. 
Consider the situation in which
the final player to arrive
has a valuation function that
only values the cake allocated
to the first player to depart. 
Whatever cake is allocated to the final player
to arrive will be of no value
to them. Hence the cake cutting
procedure cannot be proportional. 

Suppose the procedure is envy free. 
We consider the case where all players have
valuation functions that assign
some value to every slice. 
Every player is allocated some
cake otherwise they will envy
the players who are allocated cake
(and by assumption a cake cutting procedure must
allocate cake to at least one player). 
As before, the cake allocated to the
first player to depart cannot depend
on the valuation function of
the final player to arrive.
We now modify the valuation
function of the last player
to arrive so that the value of 
the cake remaining when the first player
departs is $\frac{1}{n^2}$
of the value it was before. 
Even if we allocate all the remaining cake
to the last player to arrive, the value of 
this cake cannot now equal
the value they assign to the cake
allocated to the first player to depart. 
Hence the last player to arrive will envy the 
first player to depart. 
By a similar argument, 
the procedure cannot be equitable. 
\myqed

Online cake cutting procedures
can, however, possess many of the other 
properties. 

\begin{mytheorem}
Online cake cutting procedures
can be forward proportional, 
forward envy free, weakly Pareto optimal,
truthful, scale invariant, sequential and order monotonic.
\end{mytheorem}
\myproof
Consider the online cake
cutting procedure which 
allocates all the cake to the first player 
to arrive. 
\myqed

Unfortunately, allocating all the 
cake to one player is not
very fair to the other players. 
We therefore consider some
specific online cake cutting 
procedures which divide
the cake more equitably. 
It remains an important open problem
to identify natural axioms that
these procedures satisfy which 
are not satisfied by the trivial 
allocation of all cake to one player.  

\section{Online Cut-and-Choose}

The cut-and-choose procedure for two
players dates back to antiquity. 
It appears nearly three thousand
years ago in Hesiod's poem {\em Theogeny}
where Prometheus
divides a cow and Zeus selects the
part he prefers. 
Cut-and-choose is also enshrined in the United
Nation's 1982 Convention of the Law of the Sea
where it is put forward to divide the seabed for mining. 
In cut-and-choose, one player
cuts the cake and the other takes
the ``half'' that they most prefer. 
Cut-and-choose is 
proportional, envy free, Pareto optimal, 
weakly truthful, and surjective. 
However, it
is not equitable, nor it is truthful. 

We can use cut-and-choose as the
basis of an online cake cutting procedure. 
The first player to arrive cuts
the cake and waits for the next
player to arrive. Either the next player to arrive
chooses this piece and departs, or the
next player to arrive declines this piece 
and the waiting player takes this piece and departs.
If more players are to arrive, the remaining
player cuts the cake and we repeat the process.
Otherwise, the remaining player is the
last player to be allocated cake and departs
with whatever is left. 
We assume that all players 
know how many players will arrive. 

\begin{example}
Suppose there are three players,
the first player
values only $[\frac{1}{2},1]$,
the second player
values only $[\frac{1}{3},1]$,
and the third player
values only $[0,\frac{3}{4}]$.
We suppose that they uniformly
value slices within these intervals.
If we operate the online version
of cut-and-choose, the first
player will arrive and cut off the
slice $[0,\frac{2}{3}]$ as they
assign this slice $\frac{1}{3}$ the total
value of the cake. 
The second player then arrives. 
As they assign this
slice with $\frac{1}{2}$ the total value of
the cake and they are only expecting
$\frac{1}{3}$ of the total, the second player is 
happy to take this slice and
depart. The first player then cuts off the
slice $[\frac{2}{3},\frac{5}{6}]$ as 
they assign this $\frac{1}{3}$
of the total value of the cake (and $\frac{1}{2}$
of the value remaining after the second
player departed with their slice). 
The third player then arrives.
As they assign the slice 
$[\frac{2}{3},\frac{5}{6}]$ 
with all of the 
total value of the remaining cake and
they are only expecting $\frac{1}{2}$ of 
whatever remains, the
third player is happy to
take this slice and depart.
The first player now takes
what remains, the slice 
$[\frac{5}{6},1]$. 
It can be claimed that everyone 
is happy as the first player received a ``fair''
proportion of the 
cake, whilst the other two players
received slices that were of even greater proportional
value to them. 
\end{example}

This online version of the cut-and-choose
procedure has many (but not all) of the
desirable properties described
earlier. 

\begin{mytheorem}
The online cut-and-choose procedure
is forward proportional, immediately envy free,
weakly truthful, surjective, scale invariant
and sequential. 
However, it is not
proportional, (forward) envy free, equitable,
(weakly) Pareto optimal, truthful or order monotonic. 
\end{mytheorem}
\myproof
Consider the player cutting the
cake. As they are risk averse,
and as there is a chance that they
will have to take the slice of cake
that they cut, they will cut a slice
that is at least $\frac{1}{k}$ of the
total remaining value where $k$
is the number of players still to
be allocated cake. Similarly
they will not cut a slice that is more than 
$\frac{1}{k}$ of the
total remaining value
for fear that the next player
to arrive will take it, leaving behind
cake that if it is divided proportionally
gives them a slice of small value. 
Hence, the procedure is forward proportional
and weakly truthful. 
It is also immediately envy free
since each slice that the cutting
player sees being allocated has the same value. 
To demonstrate surjectivity,
consider the partition that 
allocates the $i$th player with
the slice $[a_i,a_{i+1}]$ 
where $a_1=0$ and $a_{n+1}=1$. 
We construct a valuation 
for the $i$th player ($i<n-1$)
that assigns a value $0$ to $[0,a_i]$,
a value $1$ to $[a_i,a_{i+1}]$,
a value $0$ to $[a_{i+1},a_{i+2}]$,
a value $n-i$ to $[a_{i+2},1]$.
For the $n-1$th player,
we construct a valuation function
that assigns a value $0$ to $[0,a_{n-1}]$,
and values of $1$ to both $[a_{n-1},a_n]$
and  $[a_n,1]$. 
Finally, 
we construct a valuation function
for the $n$th player
that assigns a value $0$ to $[0,a_{n}]$,
and a value of $1$ to $[a_n,1]$. 
With these valuation
functions, the $i$th player gets the slice $[a_i,a_{i+1}]$. 
Finally, it is easy to see that the procedure is scale invariant
and sequential.

To show that this procedure is not proportional, (forward)
envy free, equitable, (weakly) Pareto optimal
truthful or order monotonic consider 4 players
and a cake in which
the first player places a value of
3 units on $[0,\frac{1}{4}]$,
1 unit on $[\frac{1}{4},\frac{3}{4}]$
and 8 units on $[\frac{3}{4},1]$, 
the second player
places a value of 
0 units on $[0,\frac{1}{4}]$,
4 units on $[\frac{1}{4},\frac{1}{2}]$,
8 units on $[\frac{1}{2},\frac{5}{8}]$,
and 0 units on $[\frac{5}{8},1]$,
the third player
places a value of 
6 units on $[0,\frac{1}{4}]$
0 units on  $[\frac{1}{4},\frac{1}{2}]$, 
1 unit on $[\frac{1}{2},\frac{5}{8}]$,
2 units on $[\frac{5}{8},\frac{3}{4}]$, 
and 3 units on $[\frac{3}{4},1]$,
and the fourth player
places a value of 
0 units on $[0,\frac{1}{4}]$,
9 units on $[\frac{1}{4},\frac{1}{2}]$, 
1 unit on $[\frac{1}{2},\frac{3}{4}]$,
and 2 units on $[\frac{3}{4},1]$.

If we apply the online cut-and-choose
procedure, the first player will 
cut off and keep the slice $[0,\frac{1}{4}]$,
the second player will cut off and keep 
$[\frac{1}{4},\frac{1}{2}]$,
The third player will now cut the
cake into two pieces: $[\frac{1}{2},\frac{3}{4}]$
and $[\frac{3}{4},1]$.
The fourth player will
take the slice $[\frac{3}{4},1]$,
leaving the third player with
the slice $[\frac{1}{2},\frac{3}{4}]$

The procedure is not proportional as the
fourth player only receives $\frac{1}{6}$ of
the total value of the cake, 
not (forward) envy free as the
first player envies the fourth player, 
and not equitable as players receive
cake of different value.
The procedure is not (weakly) Pareto optimal
as allocating the first player
with $[\frac{3}{4},1]$, the second player
with $[\frac{1}{2},\frac{3}{4}]$,
the third player with $[0,\frac{1}{4}]$,
and the fourth player with $[\frac{1}{4},\frac{1}{2}]$
gives all players a slice of greater 
value. 

The proceure is not truthful
as the second player can 
get a larger and more valuable 
slice by misrepresenting their preferences
and cutting the cake
into the slice $[\frac{1}{4},\frac{5}{8}]$.
Finally, the procedure is not order
monotonic as the value of the cake
allocated to the fourth player
decreases from 2 units to $\frac{3}{2}$
units when they arrive
before the third player. 
\myqed

\section{Online moving knife}

Another class of procedure for cutting
cakes uses one or more moving knives. 
For example, in the Dubins-Spanier procedure
for $n$ players 
\cite{dsam61},  a knife is moved across
the cake from left to right. 
When a player shouts ``stop'',
the cake is cut and this player
takes the piece to the left of the knife.
The procedure then continues with the remaining $n-1$ 
players until just one player is left
(who takes whatever remains). 
This procedure is proportional but is not
envy-free. However, only the first $n-2$ 
players to be allocated slices of cake
can be envious. 

We can use the Dubins-Spanier procedure as the
basis of an online moving knife procedure. 
The first $k$ players ($k \geq 2$) to arrive perform
one round of a moving knife procedure to select
a slice of the cake. Whoever chooses
this slice, departs. At this point, if all 
players have arrived, we continue the
moving knife procedure with $k-1$
players. Alternatively the next
player arrives and we start again 
a moving knife procedure with $k$ players. 
As before, we assume that all players 
know how many players will arrive. 

\begin{example}
Consider again the example in which 
there are three players,
the first player
values only $[\frac{1}{2},1]$,
the second player
values only $[\frac{1}{3},1]$,
and the third player
values only $[0,\frac{3}{4}]$.
If we operate the online version
of the moving knife procedure, the first
two players will arrive and 
perform one round of the moving
knife procedure. 
The second player will
be the first to call ``cut'' 
and will depart with the 
slice $[0,\frac{5}{9}]$ (as this
has $\frac{1}{3}$ of the
total value of the cake for them).
The third player will
now arrive and perform a 
round of the moving knife
procedure with the first
player using the remaining cake, $[\frac{5}{9},1]$.
The third player will be the
first to call ``cut'' and
will depart with the slice
$[\frac{5}{9},\frac{47}{72}]$
(as this has $\frac{1}{2}$ 
the total value of the remaining
value for them). 
The first player will then depart
with what remains, the slice
$[\frac{47}{72}]$. 
It can be claimed that everyone 
is happy as the second and third
players received a ``fair''' proportion of the 
cake that was left when they first arrived, whilst
the first player received an even greater 
proportional value. 
\end{example}

This online version of the moving knife
procedure has the same desirable properties 
as the online version of the cut-and-choose
procedure. 

\begin{mytheorem}
The online moving knife procedure
is forward proportional, immediately envy free,
weakly truthful, surjective, scale invariant and sequential. 
However, it is not
proportional, (forward) envy free, equitable, 
(weakly) Pareto optimal, truthful or order monotonic. 
\end{mytheorem}
\myproof
Suppose $j$ players ($j>1$) have still
to be allocated cake. Consider
any player who has arrived. They 
will call ``cut'' as soon
as the knife reaches $\frac{1}{j}$ of the value 
of the cake left for fear that
they will will receive cake of 
less value at a later stage. 
Hence, the procedure is
weakly truthful and forward proportional.
The procedure is also immediately
envy free as they will assign less value 
to any slice that is allocated
after their arrival and before their
departure. 
To demonstrate surjectivity,
consider the partition that 
allocates the $i$th player with
the slice $[a_i,a_{i+1}]$ 
where $a_1=0$ and $a_{n+1}=1$. 
We construct a valuation 
for the $i$th player ($i<n$)
that assigns a value $0$ to $[0,a_i]$,
a value $1$ to $[a_i,a_{i+1}]$,
a value $n-i$ to $[a_{i+1},1]$.
Finally, 
we construct a valuation function
for the $n$th player
that assigns a value $0$ to $[0,a_{n}]$,
and a value of $1$ to $[a_n,1]$. 
With these valuation
functions, the $i$th player gets the slice $[a_i,a_{i+1}]$. 
Finally, it is easy to see that the
procedure is scale invariant and sequential. 

To show that this procedure is not proportional, (forward)
envy free, equitable, (weakly) Pareto optimal
truthful consider
again the example with 4 players used
in the last proof. 
We suppose that $k=2$ (i.e. at any one time,
two players are watching the knife).
The first player calls ``cut'' and
departs with the slice $[0,\frac{1}{4}]$.
The second player calls ``cut'' and
departs with the slice $[\frac{1}{4},\frac{1}{2}]$.
Finally, the third player
calls ``cut'' and 
departs with the slice $[\frac{1}{2},\frac{2}{4}]$,
leaving the fourth player with
the slice $[\frac{3}{4},1]$.

The procedure is not proportional as the
fourth player only receives $\frac{1}{6}$ of
the total value of the cake, 
not (forward) envy free as the
first player envies the fourth player, 
and not equitable as players receive
cake of different value.
The procedure is not (weakly) Pareto optimal
as allocating the first player
with $[\frac{3}{4},1]$, the second player
with $[\frac{1}{2},\frac{3}{4}]$,
the third player with $[0,\frac{1}{4}]$
and the fourth player with $[\frac{1}{4},\frac{1}{2}]$
gives all players a slice of greater 
value. 

The proceure is not truthful
as the second player can 
get a larger and more valuable 
slice by misrepresenting their preferences
and not calling ``cut'' until the knife
is about to reach $\frac{5}{8}$th of the
way along the cake. 

Finally, to show that the procedure is not order
monotonic consider 3 players
and a cake in which
the first player places a value of
2 units on each of $[0,\frac{1}{3}]$,
$[\frac{1}{3},\frac{2}{3}]$,
and $[\frac{2}{3},1]$, 
the second player
places a value of 
0 units on $[0,\frac{1}{3}]$,
3 units on each of $[\frac{1}{3},\frac{2}{3}]$
and $[\frac{2}{3},1]$,
and the third player
places a value of 
2 units on $[0,\frac{1}{6}]$,
0 units on each of $[\frac{1}{6},\frac{1}{3}]$
and $[\frac{1}{3},\frac{2}{3}]$,
and 4 units on $[\frac{2}{3},1]$.
As before, we suppose that $k=2$ (i.e. at any one time,
two players are watching the knife).
The first player calls ``cut'' and
departs with the slice $[0,\frac{1}{3}]$.
The second player calls ``cut'' and
departs with the slice $[\frac{1}{3},\frac{2}{3}]$,
leaving the third player with the slice
$[\frac{2}{3},1]$. On the other hand, 
if the third player arrives ahead
of the second player then the value
of the cake allocated to them 
drops from 4 units to 2 units. 
Hence the procedure is not order
monotonic. 
\myqed

\section{Online Mark-and-Choose}

A possible drawback of both of the 
online cake cutting procedures proposed
so far is that the first player to arrive
can be the last player to 
depart. What if we want a 
procedure in which players can depart
soon after they arrive? 
The next procedure has such a property. 
Players will depart as soon as the next player arrives
(except for the last player to arrive who takes
whatever cake remains). 
However, the new procedure
is no longer sequential. It may not
allocated cake from one end. 
In addition, the new procedure
does not necessarily allocate
continuous slices of cake. 

In the online mark-and-choose procedure,
the first player
to arrive marks the cake into
$n$ pieces. The second player
to arrive selects one piece to give
to the first player who then 
departs. The second player
then marks the remaining cake into 
$n-1$ pieces and waits for the
third player to arrive. The procedure
repeats in this way until the last player
arrives. The last player to arrive selects
which of the two halves marked by the penultimate
player should be allocated to the penultimate 
player. The last player
then takes whatever remains.

\begin{example}
Consider again the example in which 
there are three players,
the first player
values only $[\frac{1}{2},1]$,
the second player
values only $[\frac{1}{3},1]$,
and the third player
values only $[0,\frac{3}{4}]$.
If we operate the online version
of the mark-and-choose procedure, the first
player will arrive and 
mark the cake into 3 equally valued 
pieces:
$[0,\frac{2}{3}]$,
$[\frac{2}{3},\frac{5}{6}]$,
and 
$[\frac{5}{6},1]$,
The second player then
arrives and selects the least valuable piece
for the first player to take.
In fact, both 
$[\frac{2}{3},\frac{5}{6}]$
and 
$[\frac{5}{6},1]$
are each worth $\frac{1}{4}$ of the
total value of the cake to the 
second player. They will
therefore choose between them
arbitrarily. Suppose
the second player decides to give the slice
$[\frac{2}{3},\frac{5}{6}]$
to the first player.
Note that the first player assigns
this slice with $\frac{1}{3}$ of the total value
of the cake. This leaves behind two sections of 
cake: $[0,\frac{2}{3}]$ and
$[\frac{5}{6},1]$.
The second player then marks
what remains into two equally valuable
pieces: the first
is the interval $[0,\frac{7}{12}]$
and the second contains
the two intervals $[\frac{7}{12},\frac{2}{3}]$
and $[\frac{5}{6},1]$. 
The third player then arrives
and selects the least valuable
piece for the second player
to take. 
The first piece is worth $\frac{7}{12}$
of the total value of the cake
to the third player. As this is over
half the total value, the other piece
must be worth less. 
In fact, the second piece is worth $\frac{1}{4}$ of
the total value. The third player 
therefore gives the second piece
to the second player. 
This leaves the third player with
the remaining slice $[0,\frac{7}{12}]$. 
It can again be claimed that everyone 
is happy as the first 
players received a ``fair''' proportion of the 
cake that was left when they arrived, whilst
both the second and third player received an even greater 
proportional value. 
\end{example}

This procedure again has the same desirable properties 
as the online version of the cut-and-choose
and moving knife
procedures. 

\begin{mytheorem}
The online mark-and-choose procedure
is forward proportional, immediately envy free,
weakly truthful, surjective, and scale invariant.
However, it is not
proportional, (forward) envy free, equitable, 
(weakly) Pareto optimal, truthful, order monotonic or 
sequential.  
\end{mytheorem}
\myproof
Any player marking the cake
will divide it into slices of equal
value (for fear that they will
be allocated one of the less valuable slices). 
Similarly, a player selecting a slice
for another player will select
the slice of least value to them (to maximize
the value that they will receive next). 
Hence, the procedure is
weakly truthful and forward proportional.
The procedure is also immediately
envy free as they will assign less value 
to the slice that they select
for the departing player than
the value of the slices that they
mark. 
To demonstrate surjectivity,
consider the partition that 
allocates the $i$th player with
the slice $[a_i,a_{i+1}]$ 
where $a_1=0$ and $a_{n+1}=1$. 
We construct a valuation 
for the $i$th player ($i<n$)
that assigns a value $0$ to $[0,a_i]$,
a value $1$ to $[a_i,a_{i+1}]$,
a value $n-i$ to $[a_{i+1},1]$.
Finally, 
we construct a valuation function
for the $n$th player
that assigns a value $0$ to $[0,a_{n}]$,
and a value of $1$ to $[a_n,1]$. 
With these valuation
functions, the $i$th player gets the slice $[a_i,a_{i+1}]$. 
Finally, it is easy to see that the
procedure is scale invariant. 

To show that this procedure is not proportional, (forward)
envy free, equitable, (weakly) Pareto optimal
or truthful consider
again the example with 4 players used
in the last two proofs. 
The first player marks and is assigned
the slice $[0,\frac{1}{4}]$ by the 
second player.
The second player then marks and is
assigned the slice $[\frac{1}{4},\frac{1}{2}]$.
The third player then marks and is
assigned the slice $[\frac{1}{2},\frac{3}{4}]$,
leaving the fourth player with the
slice $[\frac{3}{4},1]$. 

The procedure is again not proportional as the
fourth player only receives $\frac{1}{6}$ of
the total value of the cake, 
not (forward) envy free as the
first player envies the fourth player, 
and not equitable as players receive
cake of different value.
The procedure is not (weakly) Pareto optimal
as allocating the first player
with $[\frac{3}{4},1]$, the second player
with $[\frac{1}{2},\frac{3}{4}]$,
the third player with $[0,\frac{1}{4}]$,
and the fourth player with $[\frac{1}{4},\frac{1}{2}]$
gives all players a slice of greater 
value. 

The proceure is not truthful
as the second player can 
get a larger and more valuable 
slice by misrepresenting their preferences
and marking the cake into
the slices $[\frac{1}{4},\frac{5}{8}]$,
$[\frac{5}{8},\frac{3}{4}]$,
and $[\frac{3}{4},1]$. 
In this situation, the third player will allocate
the second player with the slice
$[\frac{1}{4},\frac{5}{8}]$ which is of
greater value to the second player.
It is also easy to see that the procedure is
not sequential. 

Finally, to show that the procedure is not order
monotonic consider 3 players
and a cake in which
the first player places a value of
4 units on each of $[0,\frac{1}{3}]$,
$[\frac{1}{3},\frac{2}{3}]$
and $[\frac{2}{3},1]$, 
the second player
places a value of 
0 units on $[0,\frac{1}{3}]$,
6 units on $[\frac{1}{3},\frac{2}{3}]$,
and 3 units on each of $[\frac{2}{3},\frac{5}{6}]$,
and $[\frac{5}{6},1]$, 
and the third player
places a value of 
2 unit on $[0,\frac{1}{6}]$,
0 units on each of $[\frac{1}{6},\frac{1}{3}]$
and $[\frac{1}{3},\frac{2}{3}]$, 
and 5 units on each of $[\frac{2}{3},\frac{5}{6}]$
and $[\frac{5}{6},1]$. 
The first player marks and is 
allocated the slice $[0,\frac{1}{3}]$.
The second player marks
and is allocated the slice $[\frac{1}{3},\frac{2}{3}]$,
leaving the third player with the slice
$[\frac{2}{3},1]$. On the other hand, 
suppose the third player arrives ahead
of the second player. In this 
case, the third player marks 
the cake into two slice, $[\frac{1}{3},\frac{5}{6}]$
and $[\frac{5}{6},1]$. The second player
allocates the third player the slice $[\frac{5}{6},1]$.
Hence, the value of cake allocated to the third player
drops from 10 units to 5 units when they go
second in the arrival order. 
Hence the procedure is not order
monotonic. 
\myqed

\section{Bounded number of players}

One variation of online cake cutting is when the
number of players is not known but all 
players have the (same) upper bound, $n_{max}$ on the number
of persons to be allocated cake. 
We consider three cases:
players know their arrival position 
and when the last player arrives;
players do not know their arrival
position but do know when the 
last player arrives;
players do not know when the last
player arrives. 

\subsection{Known arrival order and last player}

In this case,
each player knows how many players
have arrived before them, and players know
when no more players are to arrive. 
In this case, we can still operate the online
cut-and-choose procedure.
Given the risk
averse nature of the players, 
each player will
cut off a slice of cake of value
$\frac{1}{n_{max}-k}$ of the total where $k$ is
the number of players who
have already been allocated cake. 

\subsection{Unknown arrival order but known last player}

In this case, 
players do not know how many players
have arrived before them,
but do know when
no more players are to arrive. 
We can again operate the online
cut-and-choose procedure. 
The first player 
will cut off a slice of cake of value
$\frac{1}{n_{max}-k}$ of the total where
$k$ is the number of players already
allocated cake
 (e.g. in the first round, 
the first player cuts off a slice of value
$\frac{1}{n_{max}}$ of the total, if this 
is accepted by the second
player, they then cut off a slice
of value 
$\frac{1}{n_{max}-1}$ of the total, and so on).

We can suppose that 
the second player to arrive
will look at the cake and deduce they are the 
second player to arrive (since
they will assign the total value
of the cake to the two pieces). 
If they are not the last player
to arrive, they will accept the offered slice
if it is greater than or equal to
$\frac{1}{n_{max}}$ of the total. 
If they are the last player
to arrive, they will accept the offered slice
if it is greater than or equal to
$\frac{1}{2}$ of the total. 
Otherwise, if there
are no more players are to arrive, 
they will take
whatever cakes remain. 
If there are more players to arrive, 
they will cut off a new slice of value 
$\frac{1}{n_{max}-j}$ of the total where
$j$ is the number of players
already allocated cake
 (e.g. the second player first cuts off a slice of value
$\frac{1}{n_{max}-1}$ of the total, if this 
is accepted by the next player to arrive, the
second player then cuts off a slice
of value 
$\frac{1}{n_{max}-2}$ of the total, and so on).

We can suppose that the third (or any later) player
to arrive can only deduce that they are not
the first or second player to arrive. 
If they are not the last player to 
arrive, they will accept the offered slice
if it is greater than or equal to
$\frac{1}{n_{max}-1}$ of the total. 
If they are the last player to arrive,
they will accept the offered slice
if it is greater than or equal to
$\frac{1}{2}$ of the total. 
Otherwise, 
if there
are no more players are to arrive,  
they will take
whatever cakes remain.
If there
are more players to arrive,  
they will cut off a new slice of value 
$\frac{1}{n_{max}-j}$ of the total where
$j$ is the number of players
already allocated cake
 (e.g. they first cut off a slice of value
$\frac{1}{n_{max}-2}$ of the total, if this 
is accepted by the next player to arrive, they then cut off a slice
of value 
$\frac{1}{n_{max}-3}$ of the total, and so on).

\subsection{Unknown last player}
 
In the third case, 
players do know when
no more players are to arrive. 
We now have a potential deadlock problem
in operating the
online cut-and-choose procedure. We need
some mechanism to ensure that the
last player to arrive is allocated cake.
One option is to introduce a clock.
If a player waits longer than
a certain time, then they can take
whatever cake remains. With this 
modification, we can again operate
the online cut-and-choose procedure.

\subsection{Moving knife procedures}

We can also
use the online moving knife procedure
when there is only a bound 
on the number of players to be
allocated cake. The results
are very similar to the online
cut-and-choose procedure, and depend
on whether players know when
the last player arrives and on whether
players know how many players have
been allocated cake before them. 

\section{Related work}

There is an extensive literature on fair
division and cake cutting procedures. 
See, for instance, 
\cite{cakecut} for an
introduction. There has, however,
been considerably less work on
fair division problems similar 
to those considered here. 

Thomson considers a generalization
of fair division problems
where the number of players may increase
\cite{wmor83}.  He explores
from an axiomatic perspective
whether it is possible to have
a procedure in which players'
allocations are monotonic (i.e.
their values do not increase as
the number of players increase)
combined with other common properties like
weak Pareto optimality. 

Cloutier {\it et al.} consider
a different generalization of 
the cake cutting problem in which
the number of players is fixed
but there are multiple cakes
\cite{cmsc2010}. This can model
situations where, for example, players
wish to choose shifts across
multiple days. Note that this problem
can be reduced to multiple single
cake cutting problems unless the players'
valuations across cakes are linked
(e.g. you prefer the same shift each day compared to
different shifts). 

A number of authors have studied
distributed mechanisms for fair
division (see, for example,
\cite{cem09}).
In such mechanisms, players typically agree locally on deals to exchange
some of the goods in their possession. 
The usual goal is to identify conditions under which
the system converges to
a fair or envy free allocation.

\section{Conclusions}

We have proposed an online form of the 
cake cutting problem. 
This permits us to explore
the concept of fair division when
players arrive and depart during the
process of dividing a resource. 
It can be used to model situations,
such as on the internet, when we need
to divide resources asynchronously. 
There are many possible future directions for this
work. One extension would be to 
indivisible goods. Another extension
would be to undesirable goods (like chores)
where we want as little of them as possible. 
In addition, it would be interesting to
consider variants of the online
cake cutting problem where players
have information about the
valuation functions of the
other players. 

\section*{Acknowledgements}

{NICTA is funded by the Department of Broadband, Communications
and the Digital Economy, and the Australian Research Council.}

\bibliographystyle{plain}
\bibliography{/Users/twalsh/Documents/biblio/a-z,/Users/twalsh/Documents/biblio/pub,/Users/twalsh/Documents/biblio/a-z2,/Users/twalsh/Documents/biblio/pub2}




\begin{contact}
Toby Walsh\\
NICTA and UNSW\\
Sydney, Australia\\
\email{toby.walsh@nicta.com.au}
\end{contact}


\end{document}